\pdfoutput=1

\documentclass[10pt,twocolumn,letterpaper]{article}

\usepackage{cvpr}              % To produce the CAMERA-READY version
\usepackage{graphicx}
\usepackage{amsmath}
\usepackage{amssymb}
\usepackage{booktabs}

\usepackage{multirow}
\usepackage{xcolor}
\usepackage{amssymb}
\usepackage{bm}

\usepackage[accsupp]{axessibility}
\usepackage[pagebackref,breaklinks,colorlinks]{hyperref}

\usepackage[capitalize]{cleveref}
\crefname{section}{Sec.}{Secs.}
\Crefname{section}{Section}{Sections}
\Crefname{table}{Table}{Tables}
\crefname{table}{Tab.}{Tabs.}

\def\cvprPaperID{3344} % *** Enter the CVPR Paper ID here
\def\confName{CVPR}
\def\confYear{2022}

\begin{document}

\title{Nested Collaborative Learning for Long-Tailed Visual Recognition}

\author{Jun Li$^{1,2*}$, Zichang Tan$^{3,4}$\thanks{The first two authors contributed equally to this work}, 
Jun Wan$^{1,2}$\thanks{Corresponding author}, Zhen Lei$^{1,2,5}$, Guodong Guo$^{3,4}$\\
$^1$CBSR\&NLPR, Institute of Automation, Chinese Academy of Sciences, Beijing, China\\
$^2$School of Artificial Intelligence, University of Chinese Academy of Sciences, Beijing, China\\
$^3$Institute of Deep Learning, Baidu Research, Beijing, China\\
$^4$National Engineering Laboratory for Deep Learning Technology and Application, Beijing, China\\
%$^5$CAIR, Hong Kong Institute of Science \& Innovation, Chinese Academy of Sciences, Hong Kong\\
$^5$Centre for Artificial Intelligence and Robotics, Hong Kong Institute of Science\&Innovation, \cr Chinese Academy of Sciences, Hong Kong, China\\
{\tt\small \{lijun2021,jun.wan\}@ia.ac.cn, zlei@nlpr.ia.ac.cn, \{tanzichang, guoguodong01\}@baidu.com}
%\and
}
\maketitle
%%%%%%%%% ABSTRACT
\begin{abstract}
%Long-tailed data distribution, a common problem in practical visual recognition tasks,
%often limits the application of deep neural network in practical real-world applications.
%Despite the great success achieved by deep neural networks in recent years,
%it remains very challenging .
%Despite the great success achieved by deep neural networks in recent years,
%it still remains challenging .
%Long-tailed visual recognition has received increasing attention in recent years.
The networks trained on the long-tailed dataset vary remarkably,
despite the same training settings,
%greatly in predictions although they have been trained in same settings, 
which shows the great uncertainty in long-tailed learning.
To alleviate the uncertainty, we propose a Nested Collaborative Learning (NCL),
which tackles the problem by collaboratively learning multiple experts together. 
NCL consists of two core components, namely Nested Individual Learning (NIL)
and Nested Balanced Online Distillation (NBOD), which focus on the individual supervised learning for
each single expert and the knowledge transferring among multiple experts, respectively.
To learn representations more thoroughly, both NIL and NBOD are formulated in a nested way,
in which the learning is conducted on not just all categories from a full perspective 
but some hard categories from a partial perspective.
Regarding the learning in the partial perspective, we specifically select
the negative categories with high predicted scores as the hard categories
%the hardest categories for each sample 
by using a proposed Hard Category Mining (HCM).
%which defines the hard category as the negative category with a high predicted score.
In the NCL, the learning from two perspectives is nested, highly related and complementary,
and helps the network to capture not only global and robust features but also meticulous distinguishing ability.
%The proposed NCL concentrates on both individual learning and cooperation among multiple experts,
%and also learn from .
Moreover, self-supervision is further utilized for feature enhancement.
Extensive experiments manifest the superiority of our method with outperforming the
state-of-the-art whether by using a single model or an ensemble.
Code is available at \href{https://github.com/Bazinga699/NCL}{https://github.com/Bazinga699/NCL} 

\iffalse
In MECL, we propose a Dual Balanced Online Distillation (DBOD) to allow knowledge transferring among different experts. DBOD is different from previous distillation from two aspects:
First, DBOD is a balanced distillation, which reduces 
the long-tailed class bias in knowledge transferring.
Second, DBOD distills knowledge from two levels, 
where one concentrates on all categories for global and robust learning,
and the other focuses on some hard categories for capturing meticulous distinguishing ability.
Specifically, the hard categories is selected for each sample by using 
a proposed Hard Category Focus (HCF).
Besides the cooperation among multiple experts, 
the individual learning for each expert is also important.
We propose a Dual Individual Learning (DIL) to learn each individual expert 
under two supervisions on all categories and also hard categories.  
%The supervised loss for each individual expert also on the basis of two levels, 
%i.e., all categories and hard categories.
Moreover, self-supervision is further utilized for feature enhancement.
Extensive experiments show the superiority of our method with outperforming the
state-of-the-art whether with a single model or an ensemble.
Code will be made publicly.
\fi
\end{abstract}

%\newpage
%%%%%%%%% BODY TEXT
\section{Introduction}
\label{sec:intro}

\begin{figure}[t]
\centering
    {\includegraphics[width=1.0\linewidth]{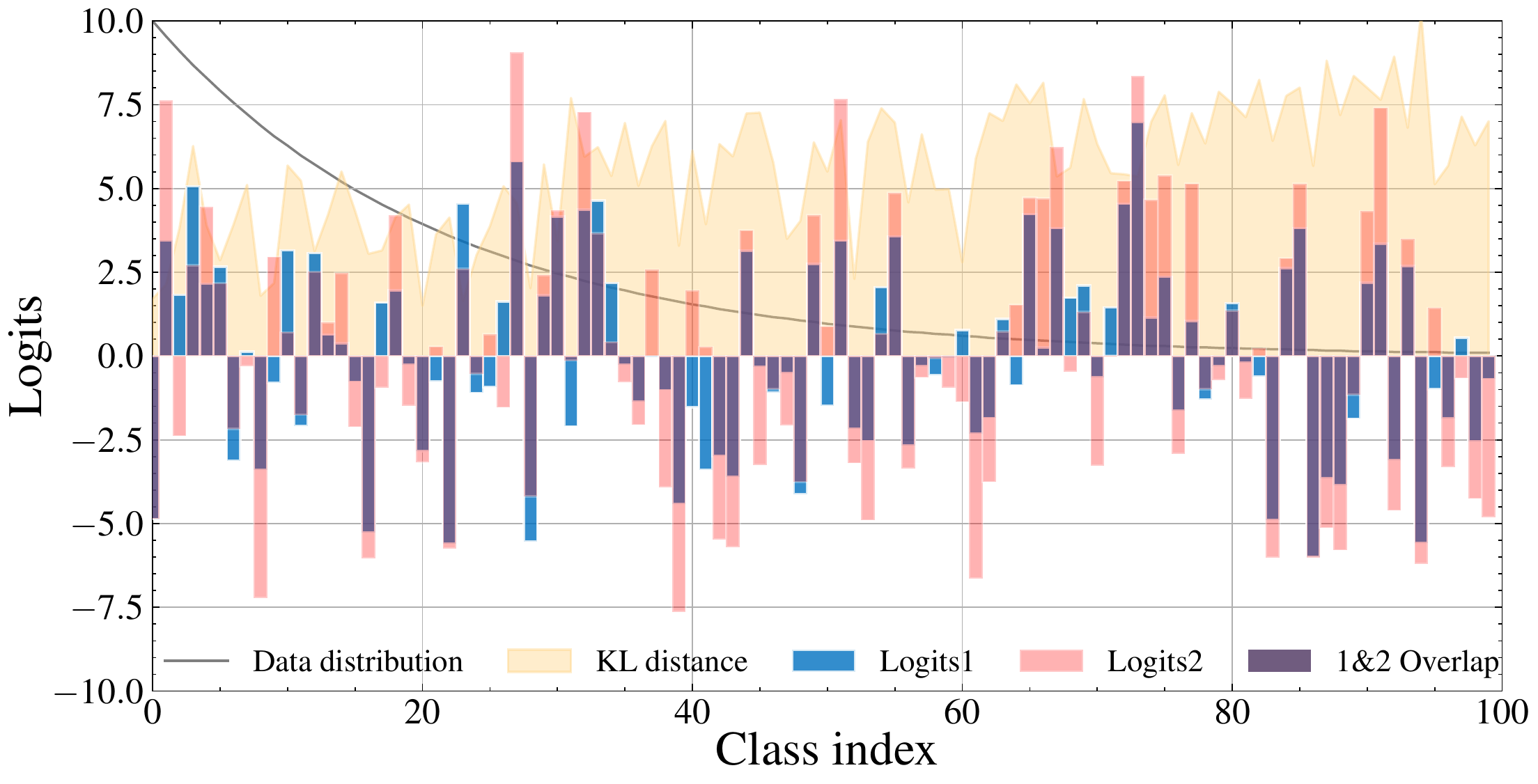}}
    \caption{ 
    The comparisons of model outputs (logits) and Kullback–Leibler (KL) distance between two networks that are trained from scratch. Analysis is conducted on CIFAR100-LT dataset with Imbalanced Factor (IF) of 100.
    The logits are visualized on the basis of a random selected example,
    and the KL distance is computed based on the whole test set and then 
    the average results of each category are counted and reported.
    Although the employed two networks have the same network structure and training settings,
    their predictions differ largely from each other especially in tail classes.
    Bested viewed in color.
}
\label{fig_intro}
\end{figure}

In recent years, deep neural networks have achieved resounding success in various visual tasks,
i.e., face analysis~\cite{wan2020multi,zhao2003face}, action and gesture recognition~\cite{Zhou_Li_Wan_2021,mitra2007gesture}.
%e.g., image classification~\cite{he2016deep,hu2018squeeze}, 
%object detection~\cite{ren2015faster,zhang2018single}, semantic segmentation~\cite{zhao2017pyramid,fu2019dual} and so on.
Despite the advances in deep technologies and computing capability,
the huge success also highly depends on large well-designed datasets of having a roughly balanced distribution,
such as ImageNet~\cite{deng2009imagenet}, MS COCO~\cite{lin2014microsoft} and Places~\cite{zhou2017places}. 
%However, those datsets are well-designed with human efforts 
%and have a roughly balanced distribution
%with each category containing adequate samples.
This differs notably
%is very different 
from real-world datasets, which usually exhibit 
long-tailed data distributions~\cite{wang2017learning,liu2019large}
where few head classes occupy most of the data while many tail classes have only few samples.
In such scenarios, the model is easily dominated by those few head classes,
whereas low accuracy rates are usually achieved for many other tail classes.
Undoubtedly, the long-tailed characteristics challenges deep visual recognition,
and also immensely hinders the practical use of deep models.

In long-tailed visual recognition, several works focus on designing the class re-balancing strategies~\cite{tan2017efficient,tan2019attention,he2009learning,cui2019class,huang2016learning,wang2017learning} and decoupled learning~\cite{kang2019decoupling,cao2019learning}.
%Those methods can accomplish some accuracy improvements but 
%still cannot deal with the long-tailed class imbalance problem well.
%For example, class re-balancing methods are often confronted with risk of overfitting.
More recent efforts aim to improve the long-tailed learning by using multiple experts~\cite{xiang2020learning,wang2020long,li2020overcoming,cai2021ace,zhang2021test}.
The multi-expert algorithms follow a straightforward idea of complementary learning,
which means that different experts focus on different aspects 
and each of them benefits from the specialization in the dominating part.
For example, LFME~\cite{xiang2020learning} formulates a network with three experts
and it forces each expert learn samples from one of head, middle and tail classes.
%Multi-expert framework usually can achieve satisfactory recognition accuracy on long-tailed recognition since each of those experts is well learned on its own field.
%We also deal with the long-tailed class distribution following 
%the multi-expert scheme due to its advantages.
Previous multi-expert methods~\cite{wang2020long,xiang2020learning,cai2021ace},
however, only force each expert to learn the knowledge in a specific area,
and there is a lack of cooperation among them.

Our motivation is inspired by a simple experiment as shown in Fig.~\ref{fig_intro},
where the different networks vary considerably, particularly in tail classes,
even if they have the same network structure and the same training settings.
This signifies the great uncertainty in the learning process.
One reliable solution to alleviate the uncertainty is 
the collaborative learning through multiple experts, namely, that each expert can be a teacher to others and also can be a student to learn additional knowledge of others.
Grounded in this, we propose a Nested Collaborative Learning (NCL) for long-tailed visual recognition.
NCL contains two main important components, namely Nested Individual Learning (NIL)
and Nested Balanced Online Distillation (NBOD),
the former of which aims to enhance the discriminative capability of each network,
and the later collaboratively transfers the knowledge among any two experts.
Both NCL and NBOD are performed in a nested way,
where the NCL or NBOD conducts the supervised learning or distillation 
from a full perspective on all categories, and also implements that 
from a partial perspective of focusing on some important categories.
Moreover, we propose a Hard Category Mining (HCM) to select the hard categories as the important categories,
in which the hard category is defined as the category that is not the ground-truth category
but with a high predicted score and easily resulting to misclassification.
The learning manners from different perspectives are nested, related and complementary,
which facilitates to the thorough representations learning.
Furthermore, inspired by self-supervised learning~\cite{he2020momentum}, 
we further employ an additional moving average model for each expert to conduct self-supervision,
which enhances the feature learning in an unsupervised manner.

In the proposed NCL,
each expert is collaboratively learned with others,
where the knowledge transferring between any two experts is allowed.
%is to improve the discriminative capability of each expert model 
%via the collaborative Learning,
NCL promotes each expert model to achieve better and even comparable performance to an ensemble's.
Thus, even if a single expert is used, it can be competent for prediction.
%only one model is also competent for forward prediction.
Our contributions can be summarized as follows:

\begin{itemize}
\setlength{\itemsep}{1.0pt}
  \item We propose a Nested Collaborative Learning (NCL) to collaboratively learn multiple experts concurrently, which allows each expert model to learn extra knowledge from others.
  %Hard Category Focus (HCF), which facilitates the network to focus on categories that are hard to be distinguished.
  \item We propose a Nested Individual Learning (NIL) and Nested Balanced Online Distillation (NBOD)
  to conduct the learning from both a full perspective on all categories and 
  a partial perspective of focusing on hard categories.
  \item We propose a Hard Category Mining (HCM) to greatly reduce the confusion with hard negative categories.
  %distinguish the sample
  %from the hard negative categories.
  %select hard categories for ,
  %and helps greatly reduce the confusion with hard categories.
  %\item We propose to enhance feature learning via self-supervision, and further employ it to assist multi-expert collaborative learning.
  \item The proposed method gains significant performance over the state-of-the-art on five popular datasets including CIFAR-10/100-LT,
  Places-LT, ImageNet-LT and iNaturalist 2018.
\end{itemize}

%and .
%Moreover, during the testing stage, a ensemble .
%and integrate all of them together during testing, 

%Existing works can be divided into three categories:
%class re-balancing～\cite{he2009learning,buda2018systematic,cui2019class,huang2016learning,ren2018learning,wang2017learning}, 
%two-stage training~\cite{kang2019decoupling,cao2019learning} and 
%multi-expert frameworks~\cite{xiang2020learning,wang2020long,li2020overcoming,cai2021ace,zhang2021test}.
%Class re-balancing is a straightforward solution for long-tailed learning, 
%and most early works follow this scheme. 
%Generally, it seeks to balance the contributions of training samples for all classes
%via data re-sampling~\cite{he2009learning,buda2018systematic} or cost-sensitive re-weighting~\cite{cui2019class,huang2016learning,ren2018learning,wang2017learning}.
%Although class re-balancing helps to ,

\section{Related Work}
\label{sec:related}

\textbf{Long-tailed visual recognition.}
To alleviate the long-tailed class imbalance,
lots of studies ~\cite{zhou2020bbn,cao2020domain,xiang2020learning,ren2020balanced,wang2020long,ye2020identifying,liu2020memory} 
are conducted in recent years. The existing methods for long-tailed visual recognition 
can be roughly divided into three categories: class re-balancing~\cite{he2009learning,buda2018systematic,cui2019class,huang2016learning,ren2018learning,wang2017learning}, 
multi-stage training~\cite{kang2019decoupling,cao2019learning} and multi-expert methods~\cite{xiang2020learning,wang2020long,li2020overcoming,cai2021ace,zhang2021test}.
Class re-balancing, which aims to re-balance the contribution of each class during training,
is a classic and  widely used method for long-tailed learning.
%Many researchers construct .
More specifically, class re-balancing consists of data re-sampling~\cite{chawla2002smote,kang2019decoupling},
%(e.g., over-sampling～\cite{chawla2002smote}, under-sampling, square-root sampling~\cite{kang2019decoupling} and progressively-balanced sampling~\cite{kang2019decoupling}),
loss re-weighting~\cite{lin2017focal,wang2021seesaw,ren2020balanced,tan2020equalization}.
%re-weighting of loss function by the numbers of different classes~\cite{lin2017focal,wang2021seesaw,ren2020balanced,tan2020equalization}.
%(e.g., Focal loss~\cite{lin2017focal}, Seesaw loss~\cite{wang2021seesaw}, BSCE~\cite{ren2020balanced} and Equalization loss~\cite{tan2020equalization}),
%and post-adjusting the model logits via label frequencies~\cite{menon2020long,zhang2021distribution}.
%(e.g.,  logit adjustment~\cite{menon2020long} and DisAlign~\cite{zhang2021distribution}).
Class re-balancing improves the overall performance but usually at the sacrifice of  the accuracy on head classes.
Multi-stage training methods divide the training process into several stages.
For example, Kang et al.~\cite{kang2019decoupling} decouple the training procedure
into representation learning and classifier learning.
Li et al.~\cite{li2021self} propose a multi-stage training strategy constructed on basis of knowledge distillation.
Besides, some other works~\cite{menon2020long,zhang2021distribution} tend to improve performance via a post-process of shifting model logits.
However, multi-stage training methods may rely on heuristic design. 
More recently, multi-expert frameworks receive increasing concern,
e.g., LFME~\cite{xiang2020learning}, BBN~\cite{zhou2020bbn}, RIDE~\cite{wang2020long}, TADE~\cite{zhang2021test} and ACE~\cite{cai2021ace}.
Multi-expert methods indeed improve the recognition accuracy for long-tailed learning, 
but those methods still need to be further exploited.
For example, most current multi-expert methods employ different models
to learn knowledge from different aspects, while the mutual supervision among them is deficient.
Moreover, they often employ an ensemble of experts to produce predictions, which leads to a complexity increase of the inference phase.

%the class re-balancing methods also can be divided into data re-sampling~\cite{he2009learning,buda2018systematic},
%cost-sensitive re-weighting~\cite{cui2019class,huang2016learning,ren2018learning,wang2017learning} 
%and logit adjustment~\cite{menon2020long,wu2021adversarial}.

\textbf{Knowledge distillation.}
Knowledge distillation is a prevalent technology in knowledge transferring.
%One typical manner of knowledge distillation is teacher-student learning~\cite{hinton2015distilling,furlanello2018born},
%which transfers knowledge from a large teacher model to a small student model.
Early methods~\cite{hinton2015distilling,passalis2018learning} often adopt an offline learning strategy,
where the distillation follows a teacher-student learning scheme~\cite{hinton2015distilling,furlanello2018born},
which transfers knowledge from a large teacher model to a small student model.
However, the teacher normally should be a complex high-capacity model and the training process may be cumbersome and time-consuming.
%with the complex high-capacity teacher and huge training time.
%which needs to train a teacher model first, and then extract the knowledge of 
%the teacher model to guide the learning of a student model.
In recent years, knowledge distillation has been extended to an online way~\cite{zhang2018deep,guo2020online,chen2020online,dvornik2019diversity},
where the whole knowledge distillation is conducted in a one-phase and end-to-end training scheme.
For example, in Deep Mutual Learning~\cite{zhang2018deep}, any one model can be a student
and can distil knowledge from all other models.
Guo et al.~\cite{guo2020online} propose to use an ensemble of soft logits to guide the learning.
Zhu et al.~\cite{lan2018knowledge} propose a multi-branch architecture with treating each branch as a student to further reduce computational cost. Online distillation is an efficient way to collaboratively learn multiple models, and facilitates the knowledge transferred among them.

\textbf{Contrastive learning.}
Many contrastive methods~\cite{he2020momentum,chen2020improved,chen2020simple,grill2020bootstrap}
are built based on the task of instance discrimination.
For example, Wu et al.~\cite{wu2018unsupervised} propose a noise contrastive estimation to compare instances
based on a memory bank of storing representations.
Representation learning for long-tailed distribution also been exploit~\cite{kang2020exploring}.
More recently, Momentum Contrast (MoCo)~\cite{he2020momentum} is proposed to produce the compared representations by a moving-averaged encoder.
%Contrastive learning often compares each sample with many negative samples to enhance the discriminative ability.
To enhance the discriminative ability,
contrastive learning often compares each sample with many negative samples.
SimCLR~\cite{chen2020simple} achieves this by using a large batch size.
Later, Chen et al.~\cite{chen2020improved} propose an improved method named MOCOv2,
which achieving promising performance without using a large batch size for training.
Considering the advantages of MoCOv2, our self-supervision is also constructed based on this structure.

\begin{figure*}[t]
\centering
    {\includegraphics[width=0.9\linewidth]{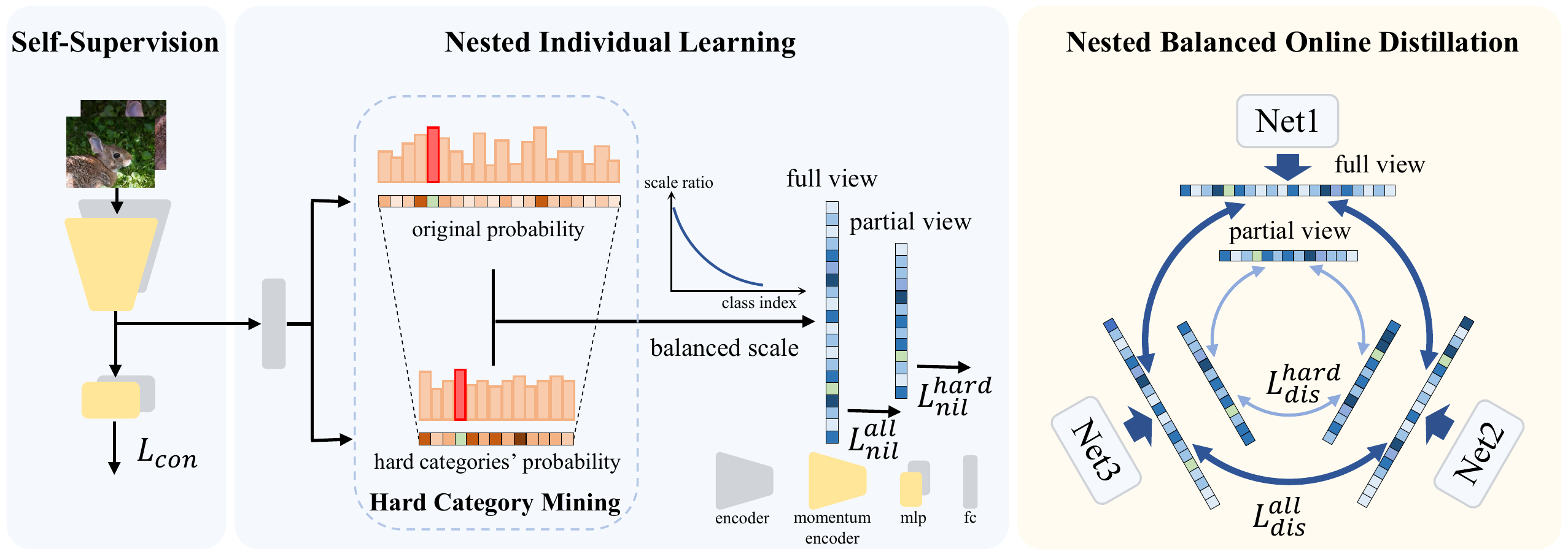}}
    \caption{
    An illustration of our proposed NCL of containing three experts. 
    The NIL enhances discriminative ability of a single expert,
    and NBOD allows knowledge transferring among multiple experts.
    NIL conducts the supervised learning from both a full 
    and a partial view, which focus on all categories and some hard categories, respectively.
    Similarly, NBOD conducts the knowledge distillation also from both a full and a partial view.
    The contrastive loss is calculated by using an extra momentum encoder and MLP layers, which can be removed in evaluation. Probabilities employed in NIL and NBOD are balanced according to the data distribution.
    }
    \label{fig_framework}
\end{figure*}

\section{Methodology}
\label{sec:method}
The proposed NCL aims to collaboratively and concurrently learn multiple experts together
as shown in Fig.~\ref{fig_framework}.
In the following, firstly, we introduce the preliminaries, 
and then present Hard Category Mining (HCM),  Nested Individual Learning (NIL),
Nested Balanced Online Distillation (NBOD) and self-supervision part.
Finally, we show the overall loss of how to aggregate them together.

\subsection{Preliminaries}
We denote the training set with $n$ samples as ${ {\mathcal D}} = \{ {\bf x}_i, y_i \}$,
where ${\bf x}_i$ indicates the $i$-th image sample and $y_i$ denotes the corresponding label.
Assume a total of $K$ experts are employed and the $k$-th expert model
is parameterized with ${\bm \theta}_k$. 
Given image ${\bf x}_i$, the predicted probability of class-$j$ in the $k$-th expert is computed as:
\begin{equation}\label{ce_prob}
     {\tilde {\bf p}}_j({\bf x}_i;{\bm \theta}_k) = \frac{exp({ {z}_{ij}^k }) }{\sum_{l=1}^C exp({{z}_{il}^k }) }
\end{equation}
where $z_{ij}^k$ is the k-th expert model's class-$j$ output and $C$ is the number of classes.
This is a widely used way to compute 
the predicted probability, and some losses like Cross Entropy (CE) loss 
is computed based on it.
However, it does not consider the data distribution,
and is not suitable for long-tailed visual recognition,
where a naive learned model based on ${\tilde {\bf p}}({\bf x}_i;{\bm \theta}_k)$
would be largely dominated by head classes.
Therefore, some researchers~\cite{ren2020balanced} proposed to 
compute predicted probability of class-$j$ in a balanced way:
\begin{equation}\label{bsce_prob}
   {\bf p}_{j} ({\bf x}_i;{\bm \theta}_k) = \frac{n_{j} exp({ {z}_{ij}^k }) }{\sum_{l=1}^C n_l exp({{z}_{il}^k  })  }  
\end{equation}
where $n_j$ is the total number of samples of class $j$.
In this way, contributions of tail classes are strengthened while contributions of head classes are suppressed.
Based on such balanced probabilities, Ren et al.~\cite{ren2020balanced}
further proposed a Balanced Softmax Cross Entropy (BSCE) loss to alleviate long-tailed class imbalance
in model training.
However, BSCE loss is still not enough, where the uncertainty in training still cannot be eliminated.

\subsection{Hard Category Mining}
In representation learning, one well-known and effective strategy to boost performance
is Hard Example Mining (HEM)~\cite{hermans2017defense}.
HEM selects hard samples for training while discarding easy samples. 
However, directly applying HEM to long-tailed visual recognition
may distort the data distribution and make it more skewed in long-tailed learning.
%HEM may also be of value to long-tailed visual recognition.
%However, HEM of discarding easy samples may distort the data distribution
%and make it more skewed in long-tailed learning.
Differing from HEM, we propose a more amicable method 
named Hard Category Mining (HCM) to exclusively select hard categories for training,
which explicitly improves the ability of distinguishing the sample from hard categories.
In HCM, the hard category means the category that is not the ground-truth category but with a high predicted score. 
Therefore, the hard categories can be selected by comparing values of model's outputs.
Specifically, we have $C$ categories in total and suppose $C_{hard}$ categories are selected to focus on.
For the sample ${\bf x}_i$ and expert $k$, the corresponding set ${\bm \Psi}_i^k$ containing the output of selected categories is denoted as:
%the model outputs ${\bm \Psi}_i$ of selected categories is denoted as:
\begin{equation}\label{hard_select}
   {\bm \Psi}_i^k = TopHard\{ z_{ij}^k | j \neq y_i \} \cup \{z_{iy_i}^k \}
\end{equation}
where $TopHard$ means selecting $C_{hard}$ examples with largest values.
In order to adapt to long-tailed learning better,
we computed the  probabilities of the selected categories in a balanced way,
which is shown as:
%Then, the classification loss like BSCE also can be employed to compute the loss 
%over the selected set ${\bm \Psi}_i$ , which facilitates the model to distinguish the input sample 
%from those confusing categories.
%Here we take HCF with the BSCE loss as an example. Then, the probabilities over the selected categories are re-computed as:
\begin{equation}\label{hard_p}
   {\bf p}^* ({\bf x}_i;{\bm \theta}_k) = \{  
   \frac{n_j exp( { {z}_{ij}^k }) }{\sum_{z_{il}^k \in {\bm \Psi}_i^k } n_l exp({{z}_{il}^k  }) }
   | {z}_{ij}^k \in {\bm \Psi}_i^k
   \}
\end{equation}

\subsection{Nested Individual Learning}
The individual supervised learning on each expert is also an important component in our NCL,
which ensures that each network can achieve the strong discrimination ability.
To learn thoroughly, we proposed a Nested Individual Learning (NIL) to perform the supervision in a nested way.
Besides the supervision on all categories for a global and robust learning,
we also force the network to focus on some important categories selected by HCM,
%and the other is the supervision only focusing on some hard categories,
which enhance model's meticulous distinguishing ability.
The supervision on all categories is trivial and constructed on BSCE loss.
Since our framework is constructed on multiple experts,
the supervision is applied to each expert and the loss on all categories over all experts is the sum of the loss of each expert:
\begin{equation}\label{nil_all}
   L_{nil}^{all} = -\textstyle{\sum\nolimits_k} log( { {\bf p}}_{y_i} ({\bf x}_i;{\bm {\bm \theta} }_k) )
\end{equation}
For the supervision on hard categories, it also can be obtained in a similar way.
Mathematically, it can be represented as:
\begin{equation}\label{nil_hard}
   L_{nil}^{hard} = -\textstyle{\sum\nolimits_k} log( { {\bf p}^*}_{y_i} ({\bf x}_i;{\bm {\bm \theta} }_k) )
\end{equation}
In the proposed NIL, the two nested supervisions are employed together to achieve a comprehensive learning,
and the summed loss is written as: 
\begin{equation}\label{loss_dil}
   L_{nil} = L_{nil}^{all} + L_{nil}^{hard}
\end{equation}

\subsection{Nested Balanced Online Distillation}
To collaboratively learn multiple experts from each other,
online distillation is employed to allow each model to learn extra knowledge from others.
Previous methods~\cite{zhang2018deep,guo2020online} consider the distillation 
from a full perspective of all categories, which aims to capture global and robust knowledge.
Different from previous methods, we propose a Nested Balanced Online Distillation (NBOD),
where the distillation is conducted not only on all categories,
but also on some hard categories
that are mined by HCM,
which facilitates the network to capture meticulous distinguishing ability.
According to previous works~\cite{zhang2018deep,guo2020online},
the Kullback Leibler (KL) divergence is employed to perform the knowledge distillation.
The distillation on all categories can be formulated as:
\begin{equation}\label{eq_dis_all}
   L_{dis}^{all} = 
   \frac{1}{K(K-1)} {\sum_{k}^{K} \sum_{q \neq k}^{K} }
   KL( {\bf p} ({\bf x}_i;{\bm \theta}_k) || {\bf p} ({\bf x}_i;{\bm \theta}_q) )
\end{equation}
As we can see, the distillation is conducted among any two experts.
Note here that we use balanced distributions instead of original distributions
to compute KL distance, which aims to eliminate the distribution bias under the long-tailed setting. 
And this is also one aspect of how we distinguish from other distillation methods.
Moreover, all experts employ the same hard categories for distillation,
and we randomly select an expert as an anchor to generate hard categories for all experts.
%And this is also the reason why we call our distillation strategy as balanced online distillation.
Similarly, the distillation on hard categories also can be formulated as:
%the other distillation on hard categories is computed as:
\begin{equation}\label{eq_dis_hard}
  L_{dis}^{hard} = 
   \frac{1}{K(K-1)} {\sum_{k}^{K} \sum_{q \neq k}^{K} }
   KL( {\bf p}^* ({\bf x}_i;{\bm \theta}_k) || {\bf p}^* ({\bf x}_i;{\bm \theta}_q) )
\end{equation}
The nested distillation on both all categories and hard categories are learned together,
which is formulated as:
\begin{equation}\label{eq_dis}
   L_{dis} = L_{dis}^{all} + L_{dis}^{hard}
\end{equation}
\iffalse
The loss $L_{kl}({\bm \theta}_k,{\bm \theta}_q)$ and $L_{kl}^*({\bm \theta}_k,{\bm \theta}_q)$
are optimized concurrently to distill the knowledge from the perspective of both all categories and hard categories.
Moreover, our framework consists of multiple experts, and the knowledge transferring 
between any two experts would be considered thoroughly.
Very natural, the proposed DBOD among all experts can be computed as:
\begin{equation}\label{mutual_all}
   L_{dis} = \frac{1}{K(K-1)} \textstyle{\sum_{k}^{K} \sum_{q \neq k}^{K} }
   \big(
   L_{kl} ({\bm \theta}_k,{\bm \theta}_q)
   + L_{kl}^*({\bm \theta}_k,{\bm \theta}_q) 
   \big)
\end{equation}
\fi

\subsection{Feature Enhancement via Self-Supervision}

Self-supervised learning aims to improve feature representations via an unsupervised manner.
Following previous works~\cite{he2020momentum,chen2020improved},
we adopt the instance discrimination as the self-supervised proxy task,
in which each image is regarded as a distinct class. 
%As shown in Fig.~\ref{*****}, 
%Following the works~\cite{he2020momentum,chen2020improved}, 
We leverage an additional temporary average model so as to conduct self-supervised learning,
and its parameters are updated following a momentum-based moving average scheme~\cite{he2020momentum,chen2020improved}
as shown in Fig.~\ref{fig_framework}.
%As shown in Fig.~\ref{fig_framework}, 
%an additional two fully connected (fc) layers and a ReLU layer are added to the backbone network 
%to generate high-level features for comparing,
%which preserves more original knowledge in the original expert model and avoids error amplification.
The employed self-supervision is also a part of our NCL,
which cooperatively learns an expert model and its moving average model to capture better features.

%For a input sample, two stochastic data augmentations is performed to generate its two views,
%and each one is input to a network. 
%and we denote their corresponding features as ${\bf z}_i$ and ${\bf z}'_i$

Take the self-supervision for expert $k$ as an example.
Let ${\bf v}_i^k$ denote the normalized embedding of the $i^{th}$ image in the original expert model,
and ${\bf {\tilde v} }_i^k$ denote the normalized embedding of its copy image with different augmentations
in the temporally average model. Besides, a dynamic queue $\mathcal{Q}^k$ is employed to 
collect historical features.
The samples in the queue are progressively replaced
with the samples in current batch enqueued and the samples in oldest batch dequeued.
Assume that the queue $\mathcal{Q}^k$ has a size of $N$ and $N$ can be set to be much larger than the typical batch size, which provides a rich set of negative samples 
and thus obtains better feature representations.
The goal of instance discrimination task is to increase the similarity of features of the same image
while reduce the similarity of the features of two different images.
We achieve this by using a contrastive learning loss, which is computed as:

\begin{small}
\begin{equation}\label{contrastive_loss}
%\small
   L_{con}^k = -log( \frac{exp( {{\bf v}_i^k}^T {\bf {\tilde v}}_i^k /\tau )}
                      { exp( { {\bf v}_i^k }^T {\bf {\tilde v}}_i^k /\tau)  +  
                      \sum_{{\bf {\tilde v}}_j^k \in { \mathcal{Q}^k}} exp( { {\bf v}_i^k }^T {\bf {\tilde v}}_j^k /\tau)  } )
\end{equation}
\end{small}
where $\tau$ is a temperature hyper-parameter.
Similar to Eq.~\ref{nil_all} and Eq.~\ref{nil_hard},
the self-supervised loss over all experts can be represented as
$L_{con} = \sum\nolimits_k L_{con}^k$.

\subsection{Model Training}

The overall loss in our proposed NCL consists of three parts:
the loss $L_{nil}$ of our NIL for learning each expert individually,
the loss $L_{dis}$ of our NBOD for cooperation among multiple experts,
and the loss $L_{con}$ of self-supervision.
%the common supervised loss to learn individual knowledge for each expert,
%including the supervised learning losses on two levels 
%($L_{bsce}$ and $L_{hcf}$ for all categories and hard categories, respectively),
%the self-supervised loss $L_{con}$ and the collaborative distillation loss $L_{dis}$.
%the one is the contrastive loss to enhance feature learning via self-supervision,
%and the another one is the distillation loss to collaboratively learn them from each other. 
The overall loss $L$ is formulated as:
\begin{equation}\label{mcl_loss}
   L =   L_{nil} + L_{con}  + \lambda L_{dis}
\end{equation}
where $\lambda$ denotes the loss weight to balance the contribution of cooperation among multiple experts.
For $L_{nil}$ and $L_{con}$, 
%these losses are designed for a singe network training,
they play their part inside the single expert,
and we equally set their weighs as 1 in consideration of generality.

\section{Experiments}

\subsection{Datasets and Protocols}
We conduct experiments on five widely used datasets,
%for long-tailed visual recognition, 
including CIFAR10-LT~\cite{cui2019class}, 
CIFAR100-LT~\cite{cui2019class}, 
ImageNet-LT~\cite{liu2019large},
Places-LT~\cite{zhou2017places},
and iNaturalist 2018~\cite{van2018inaturalist}. 
%Following the common evaluation protocol\cite{zhang2021bag,kang2019decoupling,cui2021parametric},
%the model are trained on the long-tailed training set, and evaluated on the uniform validation or test set.
%Detailed descriptions about employed datasets are clarified in the following.
%As for Protocols,  following\cite{zhang2021bag,kang2019decoupling,cui2021parametric}, we train the model on the long-tailed distribution training set and evaluate their performance on the corresponding uniform test set.

\textbf{CIFAR10-LT and CIFAR100-LT}~\cite{cui2019class} are created from 
the original balanced CIFAR datasets~\cite{krizhevsky2009learning}.
%with the number of samples decaying exponentially across different classes.
Specifically, the degree of data imbalance in datasets is controlled by an Imbalance Factor (IF),
which is defined by dividing the number of the most frequent category by that of the least frequent category.
The imbalance factors of 100 and 50 are employed in these two datasets.
\textbf{ImageNet-LT}~\cite{liu2019large} is sampled from the popular ImageNet dataset~\cite{deng2009imagenet} under long-tailed setting following the Pareto distribution with power value $\alpha$=6. 
ImageNet-LT contains 115.8K images from 1,000 categories.
%and the number of images in each category ranges from 5 to 1280.
\textbf{Places-LT} is created from the large-scale dataset Places~\cite{zhou2017places}.
This dataset contains 184.5K images from 365 categories.
%and the number of each classes ranges from 5 to 4,980.
\textbf{iNaturalist 2018}~\cite{van2018inaturalist} is the largest dataset for long-tailed visual recognition.
%The images in this dataset are collected in the wild scenarios for species identification of animals and plants.
iNaturalist 2018 contains 437.5K images from 8,142 categories, and it is extremely imbalanced with an imbalance factor of 512.

According to previous works~\cite{cui2019class,kang2019decoupling}
the top-1 accuracy is employed for evaluation.
Moreover, for iNaturalist 2018 dataset, 
we follow the works~\cite{cai2021ace,kang2019decoupling} to divide classes into many (with more than 100 images), medium (with 20 $\sim$ 100 images) and few (with less than 20 images) splits,
and further report the results on each split. 

\subsection{Implementation Details}

For CIFAR10/100-LT, following~\cite{cao2019learning, zhang2021bag}, we adopt ResNet-32~\cite{he2016deep} as our backbone network and liner classifier for all the experiments.
%Input images are randomly cropped with size 32$\times$32.
%and horizontal flipped with  the probability of 0.5. 
We utilize ResNet-50~\cite{he2016deep}, ResNeXt-50~\cite{xie2017aggregated} as our backbone network for ImageNet-LT, ResNet-50 for iNaturalist 2018 and pretrained ResNet-152 for Places-LT respectively, based on~\cite{liu2019large,kang2019decoupling,cui2021parametric}. Following~\cite{zhang2021distribution}, cosine classifier is utilized for these models. 
%Following previous works~\cite{zhou2020bbn,kang2019decoupling,cui2021parametric}, we resize the input image to 256$\times$256 pixels and take a 224$\times$224 crop from the original image or its horizontal flip. We use SGD with a momentum of 0.9 and weight decay of $2 \times 10^{-4}$ as the optimizer to train all the models. As for experiments on CIFAR, the initial learning rate is 0.1 and decreases by 0.1 at epoch 320 and 360, respectively. The learning rate for Places-LT is 0.02 and decreases by 0.1 at epoch 10 and 20. For the rest dataset, The initial learning rate is set to 0.2 and decays by a cosine scheduler to $1 \times 10^{-4}$. 
Due to the use of the self-supervision component, we use the same training strategies as PaCo~\cite{cui2021parametric}, i.e., training all the models for 400 epochs except models on Places-LT, which is 30 epochs. 
%The models of CIFAR-LT are trained on a single NVIDIA RTX3090 GPU with the batch size of 128. 
%The batch size is set to 256 during training. 
In addition, for fair comparison, following~\cite{cui2021parametric}, RandAugument~\cite{cubuk2020randaugment} is also used for all the experiments except Places-LT. The influence of RandAugument will be discussed in detail in Sec.~\ref{Component_Analysis}. 
These models are trained on 8 NVIDIA Tesla V100 GPUs.

%For the sake of generality, the loss ratio $\lambda_1$ and $\lambda_2$, which play their part inside the single expert, are set to 1. 
%The temperature $\tau$ in self-supervised loss is set to 0.2 and the size of queue N is set to 65536. 
The $\beta = C_{hard} / C$ in HCM is set to 0.3.
And the ratio of Nested Balanced Online Distillation loss $\lambda$, which plays its part among networks, is set to 0.6. The influence of $\beta$ and $\lambda$ will be discussed in detail in Sec.~\ref{Component_Analysis}.

\setlength{\tabcolsep}{5pt}
\begin{table}[t]
\centering
\resizebox{\linewidth}{!}{
\begin{tabular}{l|c|cc|cc}
\toprule[1pt]
\multirow{2}{*}{Method} & \multirow{2}{*}{Ref.} & \multicolumn{2}{c|}{ CIFAR100-LT } & \multicolumn{2}{c}{ CIFAR10-LT } \\ 
\cline{3-6} 
%\midrule[0.5pt]
 &  & 100  & 50 & 100 & 50 \\ 
 \hline
%Focal loss~\cite{lin2017focal} &  & 37.4 & 42.4 & 70.4 & 75.3\\
%Mixup~\cite{zhang2017mixup} &  & 39.5 & 45.0 & 73.1 & 77.8\\
CB Focal loss~\cite{cui2019class} & CVPR'19 & 38.7 & 46.2 & 74.6 & 79.3\\ 
LDAM+DRW~\cite{cao2019learning} & NeurIPS'19 & 42.0 & 45.1 & 77.0 & 79.3\\
LDAM+DAP~\cite{jamal2020rethinking} & CVPR'20 & 44.1 & 49.2 & 80.0 & 82.2\\
BBN~\cite{zhou2020bbn} & CVPR'20 & 39.4 & 47.0 & 79.8 & 82.2\\
LFME~\cite{xiang2020learning} & ECCV'20 & 42.3 & -- & -- & --\\
CAM~\cite{zhang2021bag} & AAAI'21 & 47.8 & 51.7 & 80.0 & 83.6\\
Logit Adj.~\cite{menon2020long} & ICLR'21 & 43.9 & -- & 77.7 & --\\
RIDE~\cite{wang2020long} & ICLR'21 & 49.1 & -- & -- & --\\
LDAM+M2m~\cite{kim2020m2m}& CVPR'21 & 43.5 & -- & 79.1 & --\\
MiSLAS~\cite{zhong2021improving} & CVPR'21 & 47.0 & 52.3 & 82.1 & 85.7 \\
LADE~\cite{hong2021disentangling} & CVPR'21 &  45.4 & 50.5 & -- & --\\
Hybrid-SC~\cite{wang2021contrastive} & CVPR'21 & 46.7 & 51.9 & 81.4 & 85.4 \\
DiVE~\cite{he2021distilling} & ICCV'21 & 45.4 & 51.3 & -- & -- \\
SSD~\cite{li2021self_iccv} & ICCV'21 & 46.0 & 50.5 & -- & --\\
ACE~\cite{cai2021ace}& ICCV'21 & 49.6 & 51.9 & 81.4 & 84.9\\
PaCo~\cite{cui2021parametric} & ICCV'21 & 52.0 & 56.0 & -- & --\\
\hline
BSCE (baseline) & -- & 50.6 & 55.0 & 84.0 & 85.8 \\
Ours (single) & -- & \textbf{53.3} & \textbf{56.8} & \textbf{84.7} & \textbf{86.8} \\
Ours (ensemble) & -- & \textbf{54.2} & \textbf{58.2} & \textbf{85.5} & \textbf{87.3} \\
\bottomrule[1pt]
\end{tabular}}\caption{Comparisons on CIFAR100-LT and CIFAR10-LT datasets with the IF of 100 and 50.
%Bold indicates the best.
}
\label{results_cifar}
\end{table}

\setlength{\tabcolsep}{6pt}
\begin{table}[t]
\centering
\resizebox{1.0\linewidth}{!}{
\begin{tabular}{l|c|cc|c}
\toprule[1pt]
\multirow{2}{*}{Method} & \multirow{2}{*}{Ref.}  & \multicolumn{2}{c|}{ImageNet-LT} &  Places-LT  \\ 
\cline{3-5} 
%\midrule[0.5pt]
 &  & Res50 & ResX50 & Res152\\ 
\hline
%Focal loss~\cite{lin2017focal} & \\
OLTR~\cite{liu2019large} & CVPR'19 & -- & -- & 35.9\\
BBN~\cite{zhou2020bbn} & CVPR'20 & 48.3 & 49.3 & --\\
%Logit Adj.~\cite{menon2020long} & \\
%\hline
%OLTR～\cite{liu2019large} & \\
NCM~\cite{kang2019decoupling} & ICLR'20 & 44.3 & 47.3 & 36.4\\
cRT~\cite{kang2019decoupling} & ICLR'20 & 47.3 & 49.6 & 36.7\\
$\tau$-norm~\cite{kang2019decoupling} & ICLR'20 & 46.7 & 49.4 & 37.9\\
LWS~\cite{kang2019decoupling} & ICLR'20 & 47.7  & 49.9 & 37.6\\
BSCE~\cite{ren2020balanced} & NeurIPS'20 & -- & -- & 38.7 \\
%LDAM+DRW~\cite{cao2019learning} & \\
%LFME~\cite{xiang2020learning} & \\
%LDAM+M2m~\cite{kim2020m2m}& \\
%CAM~\cite{zhang2021bag} & \\
RIDE~\cite{wang2020long} & ICLR'21 & 55.4 & 56.8 & --\\
DisAlign~\cite{zhang2021distribution} & CVPR'21 & 52.9 & -- & --\\
DiVE~\cite{he2021distilling} & ICCV'21 & 53.1 & -- & --\\
SSD~\cite{li2021self_iccv} & ICCV'21 &--& 56.0 & --\\
ACE~\cite{cai2021ace} & ICCV'21 & 54.7 & 56.6 & --\\
PaCo~\cite{cui2021parametric} & ICCV'21 & 57.0 & 58.2 & 41.2\\
\hline
BSCE (baseline) & -- &53.9&53.6&40.2\\
Ours (single) & -- &\textbf{57.4}& \textbf{58.4} & \textbf{41.5}\\
Ours (ensemble) & -- &\textbf{59.5}& \textbf{60.5}& \textbf{41.8}\\
\bottomrule[1pt]
\end{tabular}
}
\caption{Comparisons on ImageNet-LT and Places-LT datasets. 
%Bold indicates the best.
}
\label{results_imagenet}
\end{table}

\setlength{\tabcolsep}{5pt}
\begin{table}[t]
\centering
\resizebox{0.95\linewidth}{!}{
\begin{tabular}{l|c|ccc|c}
\toprule[1pt]
\multirow{2}{*}{Method} & \multirow{2}{*}{Ref.}  & \multicolumn{4}{c}{ iNaturalist 2018} \\ 
\cline{3-6} 
%\midrule[0.5pt]
 &   &  Many  & Medium & Few & All \\ 
 \hline
%Focal loss~\cite{lin2017focal} & \\
%\hline
OLTR~\cite{liu2019large} & CVPR'19 & 59.0 & 64.1 & 64.9 & 63.9\\
BBN~\cite{zhou2020bbn} & CVPR'20 & 49.4 & 70.8 & 65.3 & 66.3\\
DAP~\cite{jamal2020rethinking}& CVPR'20 & -- & -- & -- & 67.6\\
NCM~\cite{kang2019decoupling} & ICLR'20 & \\
cRT~\cite{kang2019decoupling} & ICLR'20 & 69.0 & 66.0 & 63.2 & 65.2\\
$\tau$-norm~\cite{kang2019decoupling} & ICLR'20 & 65.6 & 65.3 & 65.9 & 65.6\\
LWS~\cite{kang2019decoupling} & ICLR'20 & 65.0 & 66.3 & 65.5 & 65.9\\
LDAM+DRW~\cite{cao2019learning} & NeurIPS'19 & -- & -- & -- & 68.0\\
Logit Adj.~\cite{menon2020long} & ICLR'21 & -- & -- & -- & 66.4\\
%LFME~\cite{xiang2020learning} & \\
%LDAM+M2m~\cite{kim2020m2m}& \\
CAM~\cite{zhang2021bag} & AAAI'21 & -- & -- & -- & 70.9 \\
RIDE~\cite{wang2020long} & ICLR'21 & 70.9 & 72.4 & 73.1 & 72.6\\
SSD~\cite{li2021self_iccv} & ICCV'21 & \\
ACE~\cite{cai2021ace} & ICCV'21 & -- & -- & -- & 72.9\\
PaCo~\cite{cui2021parametric} & ICCV'21 & -- & -- & -- & 73.2\\
\hline
BSCE (baseline) & -- &67.5&72.0&71.5&71.6 \\
Ours (single) & -- &\textbf{72.0}&\textbf{74.9}&\textbf{73.8}&\textbf{74.2} \\
Ours(ensemble) & -- &\textbf{72.7}&\textbf{75.6}&\textbf{74.5}&\textbf{74.9} \\
\bottomrule[1pt]
\end{tabular}
}
\caption{Comparisons on iNaturalist 2018 dataset with ResNet-50. 
%Bold indicates the best.
}
\label{results_inatu}
\end{table}

\subsection{Comparisons to Prior Arts}
We compare the proposed method NCL with previous state-of-the-art methods, 
like LWS~\cite{kang2019decoupling}, ACE~\cite{cai2021ace} and so on.
Our NCL is constructed based on three experts and 
both the performance of a single expert and an ensemble of multiple experts are reported.
Besides NCL, we also report the baseline results of a network with using BSCE loss for comparisons.
Comparisons on CIFAR10/100-LT are shown in Table~\ref{results_cifar},
comparisons on ImageNet-LT and Places-LT are shown in Table~\ref{results_imagenet},
and comparisons on iNaturalist2018 are shown in Table~\ref{results_inatu}.
Our proposed method achieves the state-of-the-art performance on all datasets
whether using a single expert or an ensemble of all experts.
For only using a single expert for evaluation, 
our NCL outperforms previous methods on CIFAR10-LT, CIFAR100-LT,
ImageNet-LT, Places-LT and iNaturalist2018 with accuracies of 
84.7\% (IF of 100), 53.3\% (IF of 100), 57.4\% (with ResNet-50), 41.5\% and 74.2\%, respectively.
When further using an ensemble for evaluation, 
the performance on CIFAR10-LT, CIFAR100-LT,
ImageNet-LT, Places-LT and iNaturalist2018
can be further improved to 85.5\% (IF of 100),
54.2\% (IF of 100), 59.5\% (with ResNet-50), 41.8\%
and 74.9\%, respectively.
%For example, the performance on CIFAR100-LT (IF of 100)
%is improved from 53.3\% to 54.2\%, 
%which outperforms the previous best method PaCo by 2.2\%.
More results on many, medium and few splits are listed in \textbf{Supplementary Material}.
Some previous multi-expert methods were constructed based on a multi-branch network with higher complexity.
For example, RIDE~\cite{wang2020long} with 4 experts brings
0.4 times more computation than the original single network.
However, our method of only using a single expert for evaluation
won't bring any extra computation but still outperforms them.
Besides, despite that some previous methods employ a multi-stage training~\cite{li2021self_iccv,kang2019decoupling} 
or a post-processing~\cite{menon2020long,zhang2021distribution}
to further improve the performance, our method still outperforms them.
The significant performance over the state-of-the-art shows the effectiveness of our proposed NCL.

\subsection{Component Analysis}
\label{Component_Analysis}
\begin{figure}[t]
\centering
    {\includegraphics[width=1.0\linewidth]{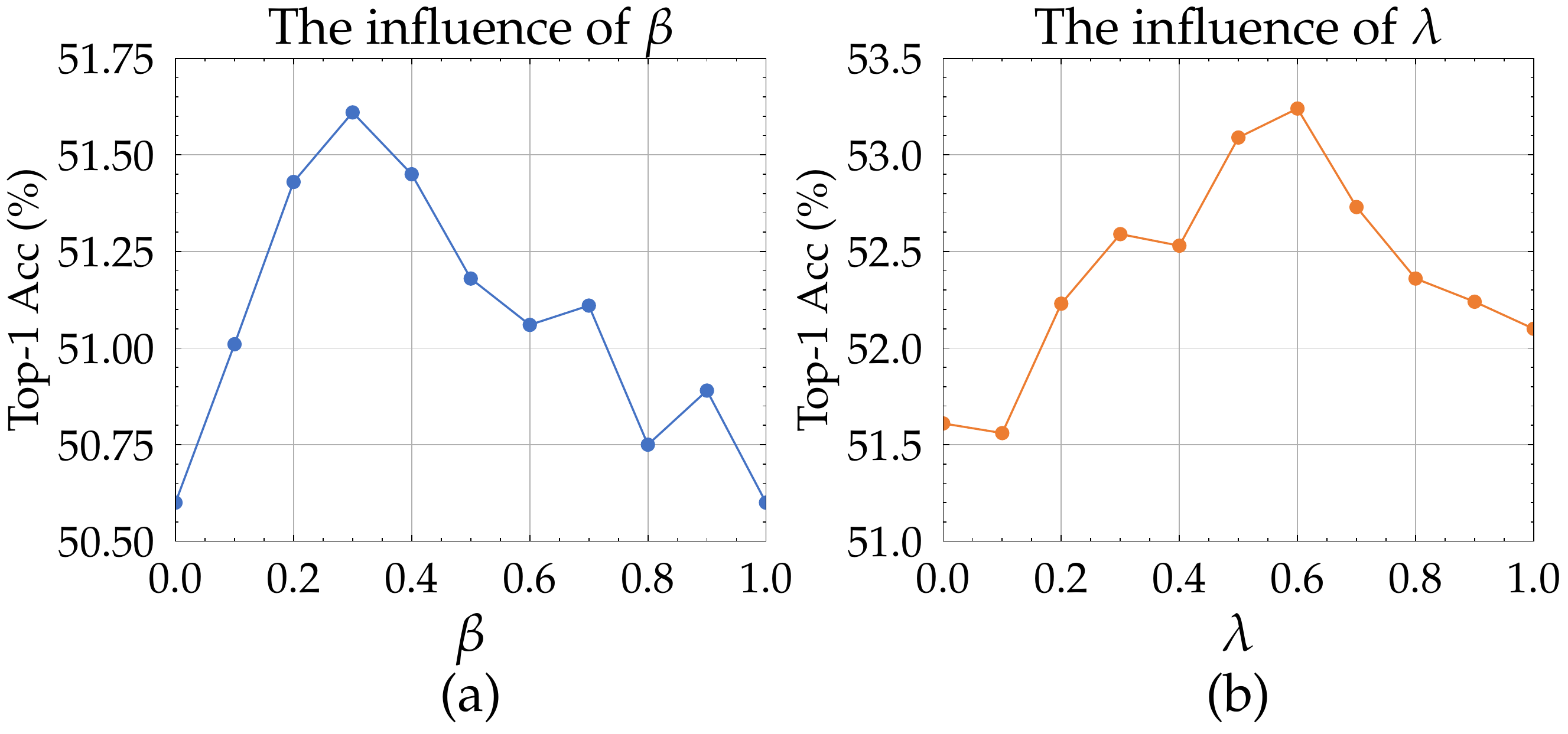}}
    \vspace{-0.6cm}
    \caption{ Parameter analysis of (a) the ratio $\beta$ and (b) the loss weight $\lambda$ on 
    CIFAR100-LT dataset with IF of 100.
    %(a) shows the experimental results of using various $\beta$
    %(a) comparisons of diferent $\beta$ on CIFAR100-LT dataset with imbalance ratio of 100. Let $\beta$ donate the value of $\frac{C_{hcf}}{C_{all}}$, the network achieves the best performance when $\beta = 0.3$, i.e., selecting 30 hard categories for CIFAR100-LT.
    }
    \label{fig_hcf}
\end{figure}

\textbf{Influence of the ratio of hard categories.}
%To enhance network's discriminating ability,
%HCF pays attention to a small number of categories that are hard to be distinguished.
The ratio of selected hard categories is defined as $\beta = C_{hard} / C$.
Experiments on our NIL model 
are conducted within the range of $\beta$ from 0 to 1 as shown in Fig.~\ref{fig_hcf} (a).
%and the corresponding experimental results are show in Fig.~\ref{fig_hcf}.
The highest performance is achieved when setting $\beta$ to 0.3.
%The performance first increases and then decreases along with the increase of $\beta$,
%and the highest performance is achieved when setting $\beta$ to 0.3.
Setting $\beta$ with a small and large values brings limited gains
due to the under and over explorations on hard categories.
%Specifically, $\beta=0.0$ means the baseline network with a BSCE loss, where HCF is not adopted. 
%The network of $\beta=1.0$ achieves similar performance with that of $\beta=0.0$,
%because $\beta=1.0$ also means classifying on all categories, 
%which is equal to the network of using BCES loss twice.

\textbf{Effect of loss weight.}
To search an appropriate value for $\lambda$,
experiments on the proposed NCL with a series of $\lambda$ 
are conducted as shown in Fig.~\ref{fig_hcf} (b).
$\lambda$ controls the contribution of 
knowledge distillation among multiple experts in total loss.
The best performance is achieved when $\lambda=0.6$,
which shows that a balance is achieved between single network training and knowledge transferring among experts.
%Intuitively, the network achieves marginal improvements with a small and a large $\lambda_1$,
%which shows that both ignoring and over emphasizing the collaborative learning are not optimal.

\begin{figure}[!t]
\centering
    {\includegraphics[width=0.8\linewidth]{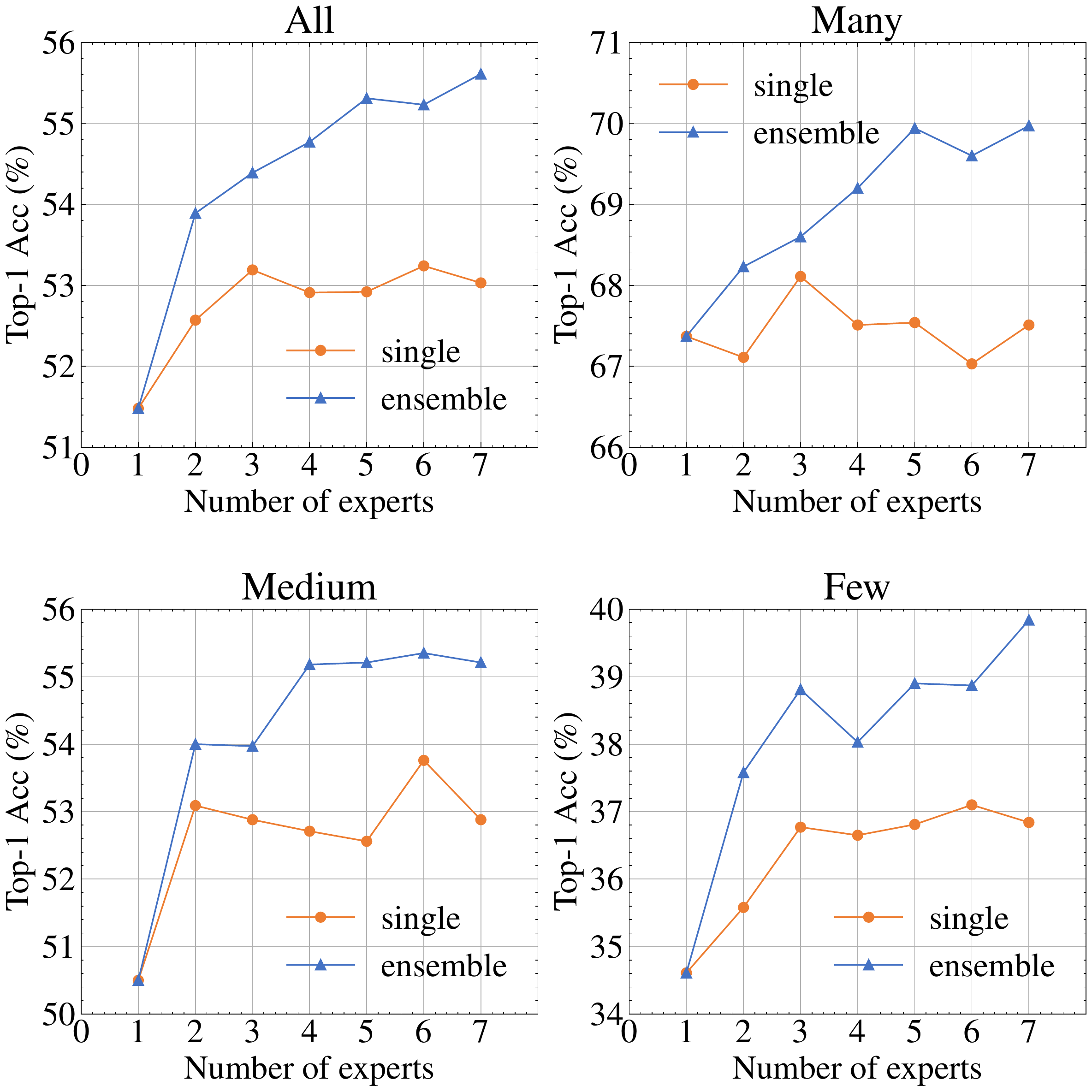}}
    \vspace{-0.3cm}
    \caption{ Comparisons of using different expert numbers on CIFAR100-LT with an IF of 100.
    We report the performance on both  a single network and an ensemble.
    Specifically, the performance on a single network is reported as the average accuracy on all experts,
    and the ensemble performance is computed based on the averaging logits over all experts.
    %'single' indicate the average valid accuracy of multi-experts, while the ensemble means the valid accuracy of average outpot logits of multi-experts.
    }
    \label{fig_multi_expert}
\end{figure}

\textbf{Impact of different number of experts.}
As shown in Fig.~\ref{fig_multi_expert}, experiments using different number of experts
are conducted. 
The ensemble performance is improved steadily as the number of experts increases,
while for only using a single expert for evaluation, 
its performance can be greatly improved when only 
using a small number of expert networks, e.g., three experts.
Therefore, three experts are mostly employed in our multi-expert framework for a balance 
between complexity and performance.

\textbf{Single expert vs. multi-expert.}
Our method is essentially a multi-expert framework,
and the comparison among using a single expert or an ensemble of multi-expert is a matter of great concern. As shown in Fig.~\ref{fig_multi_expert},
As the number of experts increases, the accuracy of the ensemble over a single expert also tends to rise.
This demonstrates the power of ensemble learning.
But for the main goal of our proposed NCL, the performance improvement over a single expert is impressive enough at the number of three.
%and the performance of using different number of experts is a matter of great concern.
%We conduct experiments with increasing the number of experts from 1 to 7,
%and experimental results are shown in Fig~\ref{fig_multi_expert}.
%Thus, we conduct experiments of various expert numbers
%and report the performance on both a singe expert and an ensemble of all experts for comparisons.
%the single network performance and ensemble performance for comparisons.
%As shown in Fig.~\ref{fig_multi_expert}, 
%the single network performance is reported when only using a expert.
%the ensemble can perform better than a single network on both all classes and many splits.
%Nevertheless, for medium and few splits, the conclusion is interesting, 
%where the single network performance is higher than that of multi-expert ensemble.
%This may be due to the fact that there is great uncertainty of predictions on medium and few splits,
%and collaborative learning is a better choice than simply averaging the logits for eliminating the uncertainty.

\textbf{Influence of data augmentations.}
Data augmentation is a common tool to improve performance. 
For example, previous works
use Mixup~\cite{zhang2021bag,cai2021ace,zhong2021improving,zhang2018mixup} %in addition PaCo~\cite{cui2021parametric} utilize 
and RandAugment~\cite{cubuk2020randaugment} to obtain richer feature representations. 
Our method follows PaCo~\cite{cui2021parametric}
to employ RandAugment~\cite{cubuk2020randaugment} for experiments.
%To investigate how much improvement this strategy can bring,
%experiments are conducted as shown in Fig.~\ref{auto_aug_analysis}.
As shown in Table~\ref{auto_aug_analysis},
the performance is improved by about 3\% to 5\% when employing RandAugment for training.
However, our high performance depends not entirely on RandAugment.
When dropping RandAugment, our ensemble model reaches an amazing performance of 49.22\%,
which achieves comparable performance to the current state-of-the-art ones.

\setlength{\tabcolsep}{8pt}
\begin{table}[t]
\centering
\resizebox{0.8\linewidth}{!}{
\begin{tabular}{c|c|c}
\toprule[1pt]
%\multicolumn{5}{c|}{ Method } & Acc \\ 
%\cline{1-5} 
%\midrule[0.5pt]
 Method    & w/o RandAug & w/ RandAug\\ 
% \midrule[0.5pt]
 \hline
CE                  &   41.88 & 44.79 \\ 
BSCE                &   45.88 & 50.60 \\
BSCE+NCL           &   47.93 & 53.31 \\
BSCE+NCL$^\dagger$ &   49.22 & 54.42 \\
\bottomrule[1pt]
\end{tabular}
}
\caption{Comparisons of training the network  with ('w/') and without ('w/o') employing RandAugment.
Experiments are conducted on CIFAR100-LT dataset with an IF of 100. $^\dagger$ Indicates the ensemble performance is reported.
%Bold indicates the best.
}
\label{auto_aug_analysis}
\end{table}

\textbf{Ablation studies on all components.}
In this sub-section, we perform detailed ablation studies for our NCL on CIFAR100-LT dataset,
which is shown in Table~\ref{analysis_ablation}.
To conduct a comprehensive analysis, 
we evaluate the proposed components including Self-Supervision ('SS' for short), 
NIL, NBOD and ensemble on two baseline settings of using CE and BSCE losses. Furthermore, for more detailed analysis, we split NBOD into two parts namely BOD$_{all}$ and BOD$_{hard}$.
%where BOD$_1$ and BOD$_2$ indicate the balanced online distillation over all categories and only hard categories, respectively.
Take the BSCE setting as an example, 
SS and NIL improve the performance by 0.82\% and 0.64\%, respectively.
And employing NBOD further improves the performance from 51.24\% to 53.19\%.
When employing an ensemble for evaluation, the accuracy is further improved and reaches the highest.
%The above analysis is constructed based on a single network, 
%and the accuracy reaches the highest when using an ensemble of all experts.
%The highest performance can be achieved when employing SS, RCF and DBOD together,
%where an accuracy of 53.31\% can be achieved.
For the CE baseline setting, similar improvements can be achieved for SS, NIL, DBOD and ensemble.
Generally, benefiting from the label distribution shift, BSCE loss can achieve better performance than CE loss.
%where the long-tailed learning has been improved by accommodating the label distribution shift.
The steadily performance improvements are achieved for all components on both baseline settings,
which shows the effectiveness of the proposed NCL.

\setlength{\tabcolsep}{3pt}
\begin{table}[t]
\centering
\resizebox{0.95\linewidth}{!}{
\begin{tabular}{ccccc|c|c}
\toprule[1pt]
%\multicolumn{5}{c|}{ Method } & Acc \\ 
%\cline{1-5} 
%\midrule[0.5pt]
NIL        & SS         & BOD$_{all}$    & BOD$_{hard}$  & Ensemble  &   Acc.@CE & Acc.@BSCE\\ 
% \midrule[0.5pt]
% \hline
% BSCE    & SS        & NIL         & BOD$_{all}$    & BOD$_{hard}$  & Ensemble  &   Accuracy \\
%         &           &             &            &            &   &44.79 \\ 
%         &           &\checkmark   &            &            &   &48.18 \\
%         &\checkmark &             &            &            &   &46.05 \\
%         &           &\checkmark   &\checkmark  &            &   &48.81 \\
%         &           &\checkmark   &\checkmark  &\checkmark  &   &49.34 \\
%         &\checkmark &\checkmark   &\checkmark  &\checkmark  &   &49.89 \\
%         &\checkmark &\checkmark   &\checkmark  &\checkmark  & \checkmark  &51.04      \\
\hline
           &             &            &            &   &44.79 &50.60 \\ 
\checkmark &             &            &            &   &48.18 &51.24 \\
           &\checkmark   &            &            &   &46.05 &51.42 \\
\checkmark &             &\checkmark  &            &   &48.81 &52.64 \\
\checkmark &             &\checkmark  &\checkmark  &   &49.34 &53.19 \\
\checkmark &\checkmark   &\checkmark  &\checkmark  &   &49.89 &53.31 \\
\checkmark &\checkmark   &\checkmark  &\checkmark  & \checkmark  &51.04 &54.42      \\
\bottomrule[1pt]
\end{tabular}
}
\vspace{-0.3cm}
\caption{Ablation studies on CIFAR100-LT dataset with an IF of 100. 'SS' indicates self-supervision.
'BOD$_{all}$' and 'BOD$_{hard}$' represent the balanced online distillation
on all categories and only hard categories, respectively.
NBOD means the setting when both 'BOD$_{all}$' and 'BOD$_{hard}$' are employed.
Experiments are conducted on the framework of containing three experts.
%$^\dagger$ indicates three experts are employed in our framework.
%Bold indicates the best.
}
\label{analysis_ablation}
\end{table}

\begin{figure}[t]
\centering
    {\includegraphics[width=1.0\linewidth]{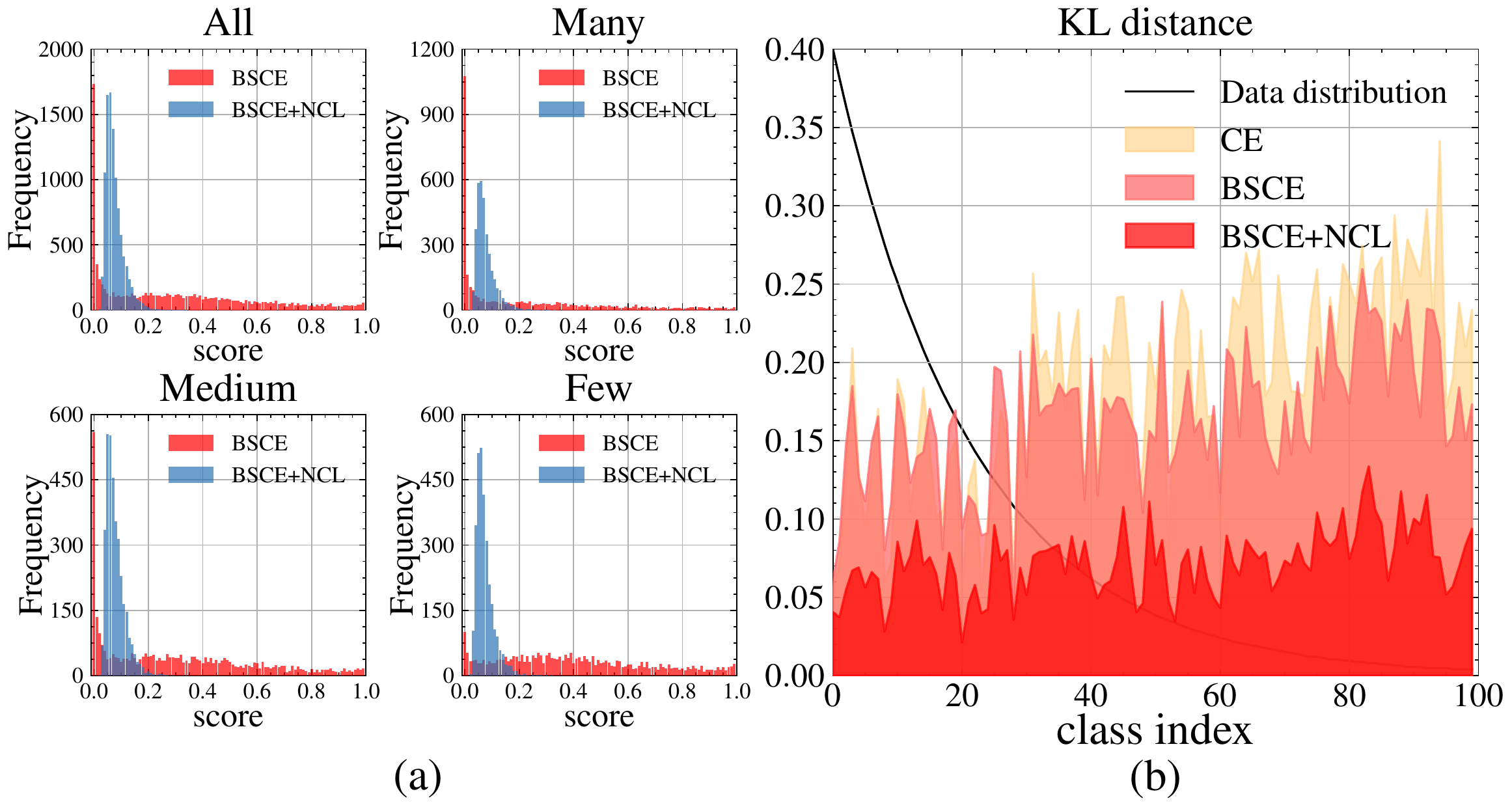}}
    \vspace{-0.8cm}
    \caption{ (a) The distribution of the largest softmax probability of hardest negative category. 
    (b) The average KL distance between two models' output probabilities on the test set. 
    %CE and BSCE indicate the two model with same structure and different  random initialization, using CE and BSCE as the supervised classification loss respectively. BSCE-MECL indicates the tow model using our proposed MECL with BSCE loss. 
    Analysis is conducted on CIFAR100-LT with an IF of 100. Best viewed in color.
    }
    \label{fig_top10}
\end{figure}

\subsection{Discussion and Further Analysis}
\iffalse
\textbf{Distribution of hard categories.}
We count the distribution of top-10 hard categories of few split during training as 
shown in Fig.~\ref{fig_top10} (a).
At the initial stage of the network training (e.g., the first ten epoches),
the network is dominated by the head classes, 
which often confuses the tail classes and results into misclassification.
With employing our method to train the network,
the frequency that the head class become a difficult category for few split is reduced 
(see the distribution in the final ten epoches).
This shows that the proposed HCF helps to reduce the long-tailed bias
and pull the sample away from the most difficult categories.
\fi

\textbf{Score distribution of hardest negative category.}
Deep models normally confuse the target sample with the hardest negative category.
Here we visualize the score distribution for the baseline method ('BSCE') 
and our method ('BSCE+NCL') as shown in Fig~\ref{fig_top10} (a).
The higher the score of the hardest negative category is, the more likely it is to produce false recognition.
The scores in our proposed method are mainly concentrated in the range of 0-0.2,
while the scores in the baseline model are distributed in the whole interval (including the interval with large values). This shows that our NCL can considerably reduce the confusion with the hardest negative category.

\textbf{KL distance of pre/post collaborative learning.}
%We visualize the average KL distance of two models trained with CE, BSCE and BSCE+NCL as shown in Fig.~\ref{fig_top10} (b).
As shown in Fig.~\ref{fig_top10} (b),
%CE and BSCE losses can hardly eliminate the uncertainty in predictions.
when networks are trained with our NCL, 
the KL distance between them is greatly reduced, which shows that 
the uncertainty in predictions is effectively alleviated.
Besides, the KL distance is more balanced than that of BSCE and CE,
which indicates that collaborative learning is of help to the long-tailed bias reduction.

\begin{figure}[t]
\centering
    {\includegraphics[width=0.8\linewidth]{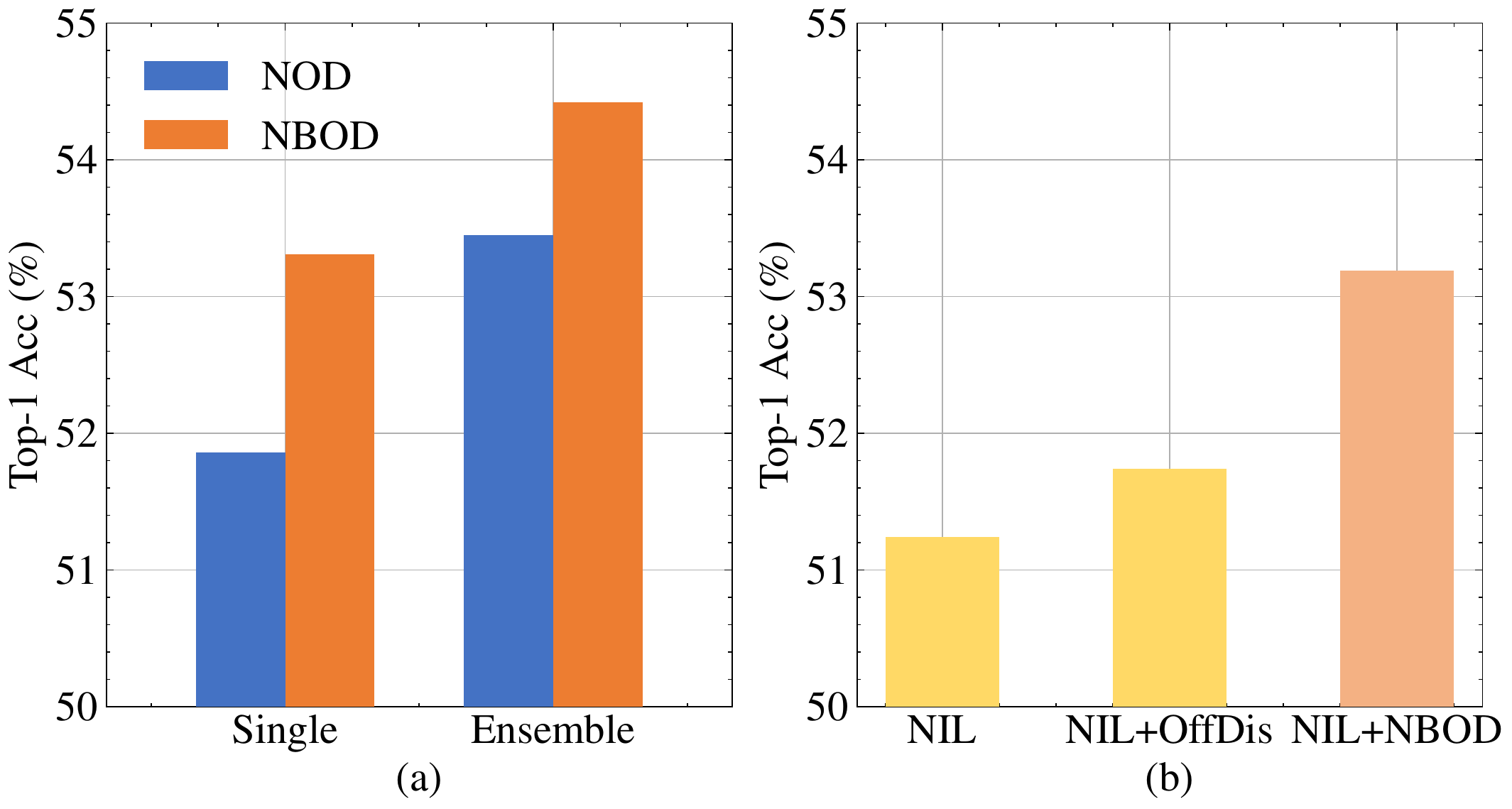}}
    \vspace{-0.3cm}
    \caption{ (a) Comparisons of using NOD or NBOD for distillation.
    (b) Comparisons of using offline distillation or our NBOD. Analysis is conducted on CIFAR100-LT with an IF of 100.
    }
    \label{balance_offdis}
\end{figure}

\textbf{NBOD without balancing probability.}
As shown in Fig.~\ref{balance_offdis} (a), 
when removing the balanced probability in NBOD (denoted as 'NOD')
both the performance of the single expert and the ensemble
decline about 1\%, which manifests the importance of employing the balanced probability
for the distillation in long-tailed learning.
%the performance of a single expert declines from 53.3\% to 52.1\%,
%and that of the ensemble also slides from 54.4\% to 53.4\%.

\textbf{Offline distillation vs. NBOD.}
To further verify the effectiveness of our NBOD, we employ an offline distillation for comparisons.
The offline distillation (denoted as 'NIL+OffDis') 
first employs three teacher networks of NIL to train individually, 
and then produces the teacher labels by using the averaging outputs over three teacher models.
The comparisons are shown in Fig.~\ref{balance_offdis} (b).
Although NIL+OffDis gains some improvements via an offline distillation, 
but its performance still 1.5\% worse than that of NIL+NBOD.
It shows that our NBOD of the collaborative learning 
can learn more knowledge than offline distillation.

\iffalse
\setlength{\tabcolsep}{6pt}
\begin{table}[t]
\centering
\resizebox{0.8\linewidth}{!}{
\begin{tabular}{c|ccccc|c}
\toprule[1pt]
%\multicolumn{5}{c|}{ Method } & Acc \\ 
%\cline{1-5} 
%\midrule[0.5pt]
 balanced probability    &  Single expert       &  Ensemble \\ 
% \midrule[0.5pt]
 \hline
            &    52.1       &  53.4     \\ 
\checkmark  &   53.3        &  54.4     \\
\bottomrule[1pt]
\end{tabular}
}
\caption{
Comparisons of with or without using the balanced probability in NBOD.
Experiments are conducted on  CIFAR100-LT dataset with an IF of 100. 
}
\label{analysis_re-balancing}
\end{table}
\fi

\section{Conclusions}
In this work, we have proposed a Nested Collaborative Learning (NCL)
to collaboratively learn multiple experts.
Two core components, i.e., NIL and NBOD, are proposed for 
individual learning of a single expert and knowledge transferring among multiple experts.
Both NIL and NBOD consider the features learning from both a full perspective and 
a partial perspective, which exhibits in a nested way. 
Moreover, we have proposed a HCM to capture hard categories for learning thoroughly.
%which is of concern when learning from a partial perspective.
%The learning on a full perspective and a partial perspective
%represents the supervisions on all categories and some hard .
%Our NCL is different from previous works in two aspects.
%First, a Hard Category Focus (HCF) has been presented to focus on some important categories 
%that are hard to distinguish rather than all categories.
%Second, a Dual Balanced Online Distillation (DBOD) has been proposed
%to exploit the cooperation complementarily and comprehensively.
Extensive experiments have verified the superiorities of our method.
%over the state-of-the-art of both a single network and an ensemble have been demonstrated.

\textbf{Limitations and Broader impacts.}
One limitation is that more GPU memory and computing power are needed when training our NCL with multiple experts.
But fortunately, one expert is also enough to achieve promising performance in inference.
%The positive impacts of this work are two-fold:
%1) it could be beneficial to bridge the gap between of research benchmarks and real world applications.
%Extensive experiments demonstrate
Moreover, the proposed method improves the accuracy and fairness of the classifier,
which promotes the visual model to be further put into practical use.
To some extent, it helps to collect large datasets without forcing class balancing preprocessing,
which improves efficiency and effectiveness of work.
%this urges us not to consider category balance when collecting large datasets,.
%One positive impact of our work is that our proposed NCL improves the accuracy and fairness of the classifier,
%which promotes the visual model to be further put into practical use.
%This will improve efficiency and effectiveness or save human costs in related areas.
The negative impacts can yet occur in some misuse scenarios, e.g., identifying minorities for malicious purposes.
Therefore, the appropriateness of the purpose of using long-tailed classification technology
is supposed to be ensured with attention.
%it is our responsibility to ensure that the purpose of using long tail classification technology is correct.

\section*{Acknowledgements}
This work was supported by the National Key Research and Development Plan under Grant 2020YFC2003901, the External cooperation key project of Chinese Academy Sciences  173211KYSB20200002, the Chinese National Natural Science Foundation Projects 61876179 and 61961160704, the Science and Technology Development Fund of Macau Project 0070/2020/AMJ,  and Open Research Projects of Zhejiang Lab No. 2021KH0AB07, and the InnoHK program.

%%%%%%%%% REFERENCES
{\small
\bibliographystyle{ieee_fullname}
\bibliography{manuscript}
}

\end{document}